\documentclass[letterpaper, 10 pt, conference]{ieeeconf}  

\IEEEoverridecommandlockouts                              
\usepackage{graphicx} 
\usepackage{xcolor}
\usepackage{hyperref}
\usepackage{cite}
\usepackage[]{mdframed}
\usepackage{booktabs} 
\usepackage{tabularx}
\usepackage{algpseudocode}

\title{Feeding the Coffee Habit: A Longitudinal Study of a Robo-Barista}
\author{Mei Yii Lim$^{1}$, David A. Robb$^{1}$, Bruce W. Wilson$^{1}$
and Helen Hastie$^{1}$
\thanks{
For the purpose of open access, the author has applied a Creative Commons Attribution (CC
BY) license to any Author Accepted Manuscript version arising.}
\thanks{$^{1}$School of Mathematical and Computer Sciences, Heriot-Watt University, EH14 4AS, UK
        {\tt\small m.lim@hw.ac.uk, d.a.robb@hw.ac.uk, bww1@hw.ac.uk, 
        h.hastie@hw.ac.uk}}%
}
\date{}

\begin{document}

\maketitle
\begin{abstract}
Studying Human-Robot Interaction over time can provide insights into what really happens when a robot becomes part of people's everyday lives. ``In the Wild" studies inform the design of social robots, such as for the service industry,  to enable them to remain engaging and useful beyond the novelty effect and initial adoption. This paper presents an ``In the Wild" experiment where we explored the evolution of interaction between users and a Robo-Barista. 
We show that perceived trust and prior attitudes are both important factors associated with the usefulness, adaptability and likeability of the Robo-Barista. A combination of interaction features and user attributes are used to predict user satisfaction. Qualitative insights illuminated users' Robo-Barista experience and contribute to a number of lessons learned for future long-term studies.
\end{abstract}

\section{Introduction}
\label{Sec: Introduction}
Social robots are expected to increasingly become part of our everyday lives. They are designed to interact socially with humans and have made appearances in various public and domestic venues including schools \cite{Kanda07}, 
malls \cite{Kanda10}, restaurants \cite{Prideaux19} and homes \cite{degraaf18}. When these robots are ensconced in our daily lives, the relationship between the human and robot will evolve over time \cite{Hinde88}. Therefore, understanding people's attitude towards, and relationship with, embedded technological artifacts will be key. As a result, it is important to understand how trust and acceptance change over long periods in human-robot interaction (HRI).  

To date, most HRI studies focus on short-term interactions between humans and robots, as conducting a longitudinal study in natural settings imposes various methodological and empirical challenges \cite{Sung09, Fernaeus10}
and requires much effort. However, such short-term studies do not generalise well to everyday life environments and interactions, within which the robots need to operate. Furthermore, robots are subject to the novelty effect - initial excitement surrounding the interaction that fades after a certain period of time. They do not capture the difficulty of sustaining engagement and high interaction quality long-term. Repeated interactions are required in many real-world robot applications for deployment (e.g. in homes, hospitals, receptions and for hospitality). This affords us the opportunity to investigate relationship-building and what leads to acceptance over a period of time.

\begin{figure}[thpb]
      \centering
      \framebox{
      \includegraphics[scale=0.6]{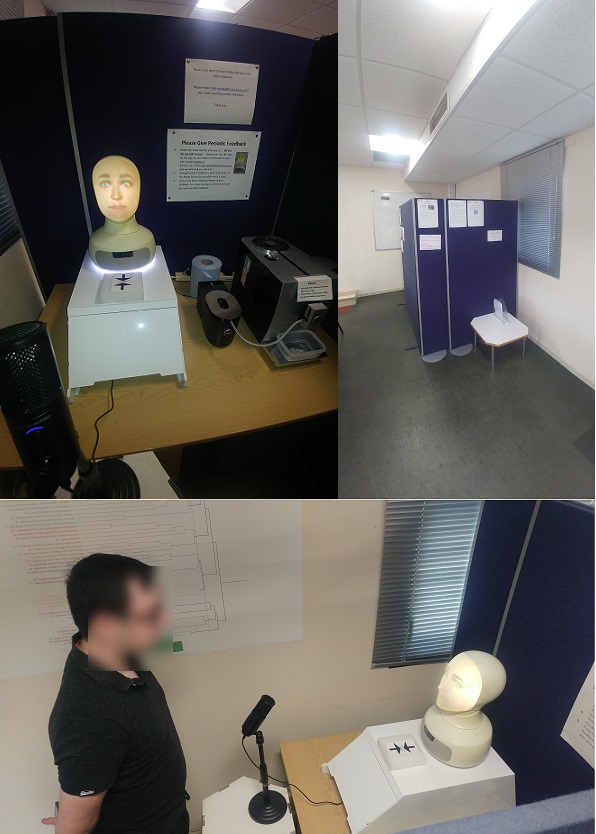}}
      \caption{The Robo-Barista installation. Top left: inside the booth
      . Top Right: the booth situated in our departmental common room. Bottom: a user interacting with the Robo-Barista}
      \label{fig:installation}
\end{figure}

In this paper, we present a longitudinal ``In the Wild" experiment with a Robo-Barista in a natural setting. A Furhat robot\footnote{https://furhatrobotics.com/ (last assessed: Mar 27, 2023)} (Figure \ref{fig:installation})  was set up in an academic department common room to interact with people and serve up the requested coffee via natural language interaction. This emulates a robot that you might find in a caf\'e, acting as a barista, with whom regular customers would interact periodically over a long period of time.

We explored the evolution of interaction between users and the Robo-Barista over 6 weeks. We focused on various social robots acceptance factors, namely perceived usefulness \cite{Heerink10}, likeability \cite{Bartneck09}, perceived trustworthiness \cite{Kraus20} and perceived adaptability \cite{Heerink10}. We took into consideration users' prior dispositions, specifically their negative attitude towards robots \cite{Nomura06} and propensity to trust robots \cite{Jessup19}, as well as usage patterns and user satisfaction. In this study, we aim to answer the following research questions:

\begin{itemize}
    \item \textbf{RQ1:} What factors play a role in the long-term user  perceptions of the Robo-Barista?
    \item \textbf{RQ2:} Did user attitude towards robots change over the study period?
    \item \textbf{RQ3:} What lessons are learned for a long-term study in the wild?
\end{itemize}

\section{Related Work}
\label{Sec: Literature Review}


Long-term effects exist in the use of technology and different interaction patterns have been observed over time and after the novelty effect wears off \cite{Koay07, Sung09}. To understand how people perceive technology, we need to study their reasons for adopting or not adopting it \cite{Young09}. Kidd and Breazeal \cite{Kidd04} posit three important factors in creating a long-term relationship between humans and robots, which we took into consideration in our study design. These are i) motivation to use the system, ii) engagement of the user, and iii) trustworthiness of the system. In order for any system to be used, users must be motivated to play an active part. The ability to draw a person into an interaction is an important first step for the user to carry out and test the abilities of a robot\cite{bohus2009}. Once interaction occurs, the user must be willing to carry on regular interactions and must have continuous belief that the robot is capable of performing its tasks over the course of the relationship.

With regards to engagement and relationship-building, we are interested in how interactions between users and the Robo-Barista evolve, the type of relationship users build and factors that are important to the relationship. According to Sung et al. \cite{Sung09}, there are 4 stages of interaction: pre-adoption, adoption, adaptation, use and retention. Based on a 6 months study of Roomba\footnote{https://www.irobot.co.uk/, last assessed: Mar 29, 2023} in 30 households, 
their study shows that users began to view the robots as social agents after the initial interaction and showed higher satisfaction in all categories of user experiences during the adoption phase. Adaptation occurs when the users make necessary changes to enable the technology to be incorporated, leading to reaffirmation or rejection of further use. 
De Graff et al. \cite{degraaf18} extended the acceptance phases to include identification, which happens 
when the technology becomes a personal object for making status claims. Their study revealed that there was an initial drop in user experience before rising again when the robot was used over a longer period of time.

In a 3 months study with severe cognitively impaired children, a seal robot PARO acted as a mediator for social interaction and was able to drive changes in people's behaviour \cite{Marti05}. 
The interaction was successful because people viewed the robots as social agents with affection rather than a mere object. 
Several studies observed decreased interaction over time due to unmet initial expectations such as the case with Pleo\footnote{https://robots.ieee.org/robots/pleo/, last assessed: Mar 29, 2023} \cite{Fernaeus10} and novelty effects wearing off \cite{Tanaka06}. Although Tanaka et al. \cite{Tanaka06} observed a lost of interest in the robot QRIO\footnote{https://robots.ieee.org/robots/qrio/, last assessed: Mar 29, 2023} over time, they also witnessed an improvement in interaction quality towards the end of their study when children hugged and pulled the robot's hand asking it to go for a walk with them. A similar pattern was observed with a dancing Robovie, despite decreased interest, children collectively created an information board and shared their knowledge about the robot towards the end of the study \cite{Kanda07}.  

In the service industry context, Berardinis et al. \cite{Berardinis20} applied different supervised learning techniques for coffee recommendations based on users’ preferences. 
Recently, Irfan et al. \cite{Irfan21} carried out a real-world study of fully autonomous interaction, evaluating the potential of data-driven approaches in generic and personalised long-term HRI. They generated two datasets\footnote{https://github.com/birfan/BaristaDatasets, last assessed: Mar 29, 2023} for a barista, one containing generic interactions between customers and a barista and one with personalised long-term interaction, where the barista would learn and recognise the users and recall their preferences. The authors argue that long-term interactions require remembering users and their preferences continuously, for a personalised experience.

Rossi and Rossi \cite{Rossi21} also stressed that effective customer retention is possible only when the robot is able to personalise the interaction to the user's needs. They found that customers' attitude towards the robot positively or negatively affects their trust and satisfaction in the service. Earlier studies have also shown that incremental learning and adaptation are vital for life-like HRI \cite{Castellano08, Lim11}

Besides user engagement, trustworthiness is key to the acceptance of the robots \cite{Kidd04,Lee13}.
A robot must be seen as dependable and trustworthy for a person to use or collaborate with it for completing a task. Trust formation begins prior to the interaction with an automated system and different layers of trust are built up sequentially \cite{Kraus20}. Hancock et al. \cite{Hancock21} found that robot-related, human-related and contextual factors are predictors of trust.  Inappropriate levels of robot trust could not only result in a frustrating experience but misuse (overly high trust) or disuse (overly low levels of trust) of the robot \cite{Lee04}.

\section{Setup}
\label{Sec: Setup}

\begin{figure}[htbp]
\centerline{\includegraphics[scale=0.5]{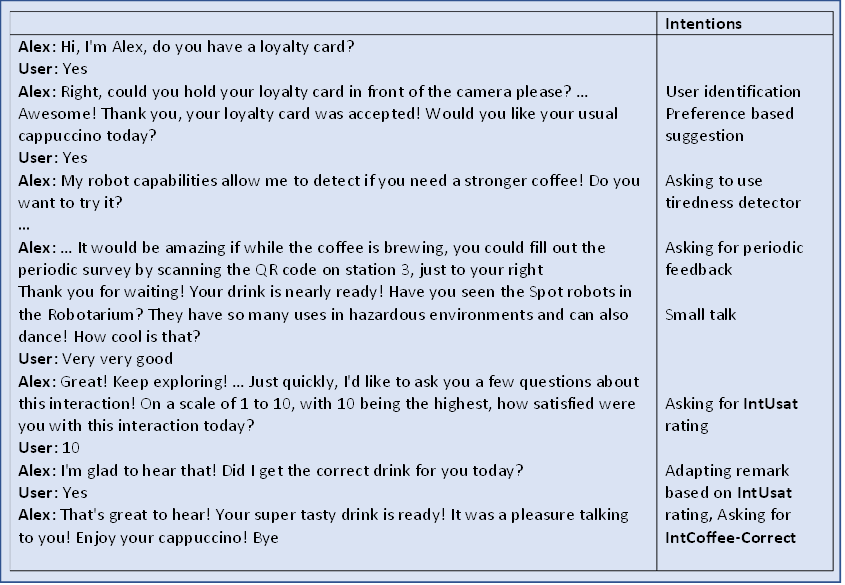}}
\caption{An excerpt of the conversation between a user and the Robo-Barista}
\label{fig:Conversation}
\end{figure}

The Robo-Barista was located at a departmental Common Room frequented by staff and postgraduate students. Its setup consists of a Furhat robot, linked up to a high-end coffee machine, Jura, via Bluetooth. The interaction starts when the Robo-Barista detects and greets a user. If the user is a first timer, it refers them to the registration signage, otherwise it asks to scan their loyalty card using its QR code scanning capability. After identifying the user, it makes order suggestions based on the user's past preferences (refer to Figure \ref{fig:Conversation} for an example) or if the user is ordering for the first time, it asks for an order, which can be a type of coffee or tea. It then confirms the request and asks if the user would like it to detect whether a stronger coffee is needed. If the user agrees, it uses its tiredness detector component developed by Cisco\footnote{https://www.cisco.com, last assessed: Mar 27, 2023} - a vision component based on Google Cloud Vision API and the Microsoft Azure Face API - to detect the user's fatigue level from their facial features. If the level is high, it suggests an extra shot of coffee \cite{lim2022demonstration}. 

The verbal interaction is implemented using an open-source conversational AI platform, RASA\footnote{https://rasa.com/, last assessed: Mar 27, 2023}. The DIET model was employed \cite{bunk2020diet} for Natural Language Understanding. It handles both user intent classification and entity recognition in a single transformer architecture and allows for the use of pre-trained embedding models, such as BERT \cite{Bert19}. For the dialogue management, the RASA TED Policy \cite{vlasov2019dialogue} is used in combination with a set of dialogue handwritten rules. The TED Policy applies a machine learning policy that can generalise patterns from example conversations and uses context from previous user utterances to perform well in a non-linear conversation. 

User identification is important for relationship establishment \cite{Hinde88}.
Unfortunately, current visual and auditory sensing technologies cannot identify users with certainty. 
To combat this problem, we introduced the loyalty card scheme, which is not uncommon in a caf\'e setting, to accurately identify a user for both data recording and interaction management.

While user questionnaires for study recruitment and exit were prompted by signage (see Section \ref{Sec: Experiment}), periodic  user feedback was requested by the robot. To reduce participation burden this was collected periodically. Specifically, for every third interaction with each user, it asks them to fill in a short feedback questionnaire before going into idle mode. Once the order is ready, the Robo-Barista informs the user and engages them in small talk (Fig. \ref{fig:Conversation}), including various topics such as weather, jokes and hobbies. It then asks the user to rate their satisfaction level on the interaction and whether they received the drink they ordered before ending the conversation.

As mentioned in Section \ref{Sec: Literature Review}, adaptation is important for personalised interaction. The Robo-Barista remembers users' previous interactions and makes suggestions for orders based on their most frequent choices. Additionally, it exhibits shared attention with the user through non-verbal behaviour and adapts the small talk to each user avoiding topics repetition. It also adapts its remarks based on the user's satisfaction response, for example, ``I am glad to hear that" when the user gave a rating above average.

\section{Experiment}
\label{Sec: Experiment}
The main aim of the experiment is to explore factors that lead to the acceptance of the Robo-Barista. We investigate how various factors change over time and if there are correlations between them. These factors are: prior attitude towards robots, perceived usefulness, likeability, perceived trustworthiness, perceived adaptability, user satisfaction and usage pattern. Throughout the following sections, we present lessons learned as they emerge from the discussion. 

Before we begin the detailed description of the study, a brief word about nomenclature: As the different collection instruments are described and measures are introduced we label the measures with the concepts that they measure. Later, to help distinguish at what stage a measure is collected, we adopt a policy of prefixing our measure labels with the study stage at which they were collected. For example, \textbf{RecruitAge} is collected on recruitment.

\begin{figure}[htbp]
\centerline{\includegraphics[scale=0.52]{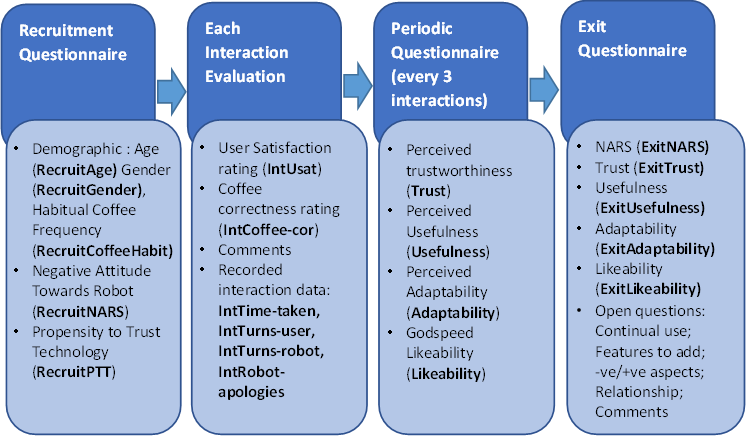}}
\caption{Questionnaires and user feedback flow}
\label{fig:feedbackflow}
\end{figure}

The procedure was ethically approved by Heriot-Watt University School of Mathematical and Computer Sciences' ethics board. The Robo-Barista 
was situated in the School Common Room. Users were recruited through signage placed beside the Robo-Barista. In order to participate, they were required to bring their own mug and have an internet connected smart phone so that they could access the online questionnaires (all questionnaires were accessible via QR codes). To sign up, they were given a loyalty card with a unique customer number as identification. These cards were in envelopes and could be picked up from a dispenser marked ``Customer Cards" below the signage. Figure \ref{fig:feedbackflow} shows the various input and feedback we elicited from the users throughout the interaction period. Table \ref{tab:MeasuresQuaire} gives the descriptive statistics for all the continuous measures (descriptions to follow) captured in Figure \ref{fig:feedbackflow}. 

 \begin{table}[th]
\centering
    \caption{Descriptive statistics for all numeric measures (Recruit and Exit n=21, Interactions (Int) n=112). IntTime-taken  is in mins:secs.}
     
     \label{tab:MeasuresQuaire}
    \begin{tabular}{{l l c c c c c}}\toprule
         \textbf{Measure} & \textbf{Mean} & \textbf{Med} & \textbf{SD} & \textbf{Max} & \textbf{Min} \\ 
        \midrule
    \small RecruitAge  & 35.6 & 34.5 & 13.50 & 23.5 & 74.5 \\
    \small RecruitNARS  & 2.39 & 2.40 & 0.48 & 1.67 & 3.47 \\
    \small RecruitPTT  & 3.42 & 3.33 & 0.65 & 2.33 & 4.50 \\ \hline
     \small ExitNARS  & 2.40 & 2.67 & 0.43 & 1.67 & 3.13 \\
    \small ExitTrust  & 4.88 & 5.00 & 0.87 & 3.00 & 6.14\\
    \small ExitUsefulness  & 4.20 & 4.33 & 1.64 & 1.00 & 6.67 \\
    \small ExitAdaptability & 4.28 & 4.33 & 0.94 & 2.67 & 6.00 \\
     \small ExitLikeability  & 63.3 & 61.8 & 15.2 & 32.0 & 88.6 \\ \hline
         \small IntUsat & 7.25 & 7.00 & 2.33 & 1 & 10 \\
    \small IntTime-taken  & 2:38 & 2:33 & 0:34 & 1:35 & 4:50 \\
    \small IntTurns-user  & 9.00 & 8.00 & 2.77 & 5 & 21 \\
    \small IntTurns-robot  & 11.14 & 11.00 & 3.09 & 6 & 25  \\
    \small IntRobot-apologies  & 1.10 & 1.00 & 1.53 & 0 & 10 \\ 
        \bottomrule
     \end{tabular}
\end{table}

\subsection{Recruitment Questionnaire}
\label{RecruitmentQuestionnaire}
After giving consent for their participation at the start of the `Recruitment Questionnaire', the users entered their customer number and some demographic information about themselves specifically: \textbf{Age Group} (18-29, 30-39, 40-49, up to 80-89 then 90 or above); \textbf{Gender} (Male, Female, Non-binary$/$third gender, Prefer not to say, Other); and their \textbf{Coffee drinking habits}  (How often do you drink coffee?: More than 1 cup per day, A cup per day, A few cups per week but not daily, None at all). They were also asked about their prior dispositions towards robots using the following measures: 
\begin{itemize}
    \item 14-item Negative Attitude Towards Robots (\textbf{NARS}) questionnaire \cite{Nomura06}
    , and
    \item 6-item Propensity to Trust Technology (\textbf{PTT}) questionnaire \cite{Jessup19}, which measures stable characteristics in individuals, attitudes towards technology and whether people were likely to collaborate with technology.
\end{itemize}

Both questionnaires use a 5-point Likert scale ranging from strongly disagree (1) to strongly agree (5). Prior work has shown that adapting the \textbf{PTT} to interaction context enhances the reliability of the measure, and hence the predictability of perceived trustworthiness \cite{Jessup19}. Therefore, we adapted the \textbf{PTT} questionnaire to our scenario with the word ``technology" being replaced by ``robot".  We also added an additional question to the \textbf{NARS} specific to our scenario - ``I would feel uneasy if I had to order a drink from robots". 

Once the users signed up, they could start ordering drinks from the Robo-Barista. They were being offered up to one cup of free coffee or tea a day. The interaction is as described in Section \ref{Sec: Setup}. Audio, but not video, was recorded throughout the interaction. 

\subsection{Each Interaction Evaluation}
\label{EachInteraction}
After each Interaction, the users provided a self report of a) their satisfaction rating (\textbf{IntUsat}) to the question ``On a scale from 1 to 10, with 10 being the highest, how satisfied with this interaction were you today?"; and b) whether they received the correct coffee (\textbf{IntCoffee-Correct}): True, False, Walked-Away (before being asked or answering), Ask-Repeat (Asked robot to repeat the question). In addition, we recorded the following interaction data: Interaction duration (\textbf{IntTime-taken}), number of user and robot turns (\textbf{IntTurns-user} and \textbf{IntTurns-robot)} and number of robot apologies (\textbf{IntRobot-apologies}). 

\begin{mdframed}
\textbf{Lesson 1: Don't expect users to patiently complete interactions}
In the wild, users can get interrupted, leave unexpectedly, and do not necessarily hang around politely until the robot has completed every step of its programmed interaction.
\end{mdframed}

\subsection{Periodic Questionnaire}
\label{PeriodicQuestionnaire}
Users were prompted, after every three coffees, to give feedback on their experience through the `Periodic Questionnaire', where they rated the following:
\begin{itemize}
    \item Perceived trustworthiness - their level of trust in the Robo-Barista using an adapted LETRAS-G (\textbf{Trust})\cite{Kraus20}, a subset of the empirically determined scale of trust in automated systems \cite{Jian00},
    \item Perceived usefulness (\textbf{Usefulness}) of the Robo-Barista \cite{Heerink10}
    \item Perceived Adaptability (\textbf{Adaptability}) of the Robo-Barista \cite{Heerink10} and
    \item Likeability from the portion of the Godspeed questionnaire (\textbf{Likeability}) \cite{Bartneck09}
\end{itemize}

The \textbf{Trust}, \textbf{Adaptability} (e.g. ``I think the robot will only do what I need at that particular moment") 
and \textbf{Usefulness} items were rated using 7-point Likert scales ranging from Strongly Disagree (1) to Strongly Agree (7). Likert scales allow only integer responses and may fail to capture the subtle effects evoked by individual variability \cite{Imbault18}. In order to allow for a more fine-grained measure and more robust detection of subtle individual differences
, sliders ranging from 0 to 100 were used for the rating of the \textbf{Likeability} items.
Additionally, a general open question, ``Any other comments?", was asked allowing the users to give worded feedback on aspects or issues.

\subsection{Exit Questionnaire}
\label{ExitQuestionnaire}

At the end of the study, users were asked to fill out an `Exit Questionnaire'.
After entering their customer number, users were given the \textbf{NARS}, \textbf{Trust}, \textbf{Usefulness}, \textbf{Adaptability} and \textbf{Likeability} items to rate. Finally, the open questions listed below were asked to help us understand better users' perception of the Robo-Barista and provide guidelines for future refinements. 

\begin{enumerate}
    \item Would you continue using the Robo-Barista? Why? 
    \item If you have the opportunity, which features would you add to the Robo-Barista?
    \item Which aspects (negative/positive) regarding the Robo-Barista could you highlight? 
    \item What is the Robo-Barista to you? (A tool to perform task, A social mediator among common room visitors, A social agent or A status symbol \cite{Sung09,degraaf18})
    \item Any other comments? 
\end{enumerate}

\section{Data Collected}

In this section, we describe the actual data collected, detailing how the numbers of responses, 
 missing values and the need to recode some categorical fields
, led to the final populations of responses included in analysis. 

On recruitment, 53 users gave consent and provided the information detailed in Subsection \ref{RecruitmentQuestionnaire}.
During the study, 186 interactions 
were recorded, 136 included responses to the in-dialogue questions for user satisfaction (\textbf{IntUsat}) and task performance (\textbf{IntCoffee-Correct}). Of those 136, 8 were by users, who had not completed the online recruitment form (i.e. no consent) and were set aside. 
Only 12 `Periodic Questionnaire' 
responses were recorded from just 8 users (one without providing consent). 

Furthermore, only 22 Exit responses were collected. One 
user had not recorded their consent (thus the response was set aside). This left 21 valid responses, where participants did both the recruitment and exit surveys (referred to as \textbf{Thru-Study} users). Of the 21 Thru-Study users, 15 identified as Male, 4 as Female, 1 as non-binary, and 1 preferred not to say. This was representative of the Common Room population during this early post-pandemic period.

We decided that, while the Recruitment and Exit returns of the 21 \textbf{Thru-Study} users could give a useful picture of the start and end of the study, the low number of Periodic returns were too few to usefully characterise Robo-Barista use during the course of the study. However, the interaction data itself was numerous enough and this is what we analyse to explore what transpired between Recruitment and Exit.


Of the 21 \textbf{Thru-Study} users, one had no fully recorded interactions with a \textbf{IntUsat} rating, four had only one, and 16 had two or more interactions. An average (\textbf{IntMeanUsat})  was calculated for users with at least one user satisfaction rating, else \textbf{IntMeanUsat} was coded as missing.

\begin{mdframed}
\textbf{Lesson 2: Unexpected use patterns}
People may use your installation in ways you had not expected: e.g. We had some users who used coffee ``loyalty" cards to access the robot and get coffee but who did not visit the online recruitment form to sign up and give consent, as per instructions on the installation signage.
\end{mdframed}

\begin{mdframed}
\textbf{Lesson 3: Attrition Rate}
Be prepared for users to drop out from an ``In the Wild'' study. 
Motivation is needed to prompt better return rates for periodic feedback (e.g. extra cups of coffee). 

\end{mdframed}


\section{Data Analysis Method} 
\label{subsec:AnalysisMethods}
The experiment was essentially an observational study, rather than variable manipulation. 
Users interacted with the Robo-Barista. We asked them questions, and recorded their interactions and responses. 
Our methods, described below, were chosen to exploit the observational nature of the data.
\subsection{Correlation}
We were able to explore possible associations between the measures collected in the study through bi-variate correlation \cite{Field09}. In Section \ref{Sec: Result and Discussion}, we report statistically significant correlations between various factors, such as users' PTT and the likeability of the robot. 
\subsection{Multiple Linear Regression (MLR)}
We probe the influences on users' in-dialogue self-reported satisfaction (\textbf{IntUsat}) using Univariate Multiple Linear Regression (MLR). This helps to understand what combination of interaction variables contribute to user satisfaction, inspired by the PARADISE framework \cite{Walker97}.  To allow a regression analysis, the variables need to be either numeric or categorical but with only two categories \cite{Field09}. 

On initial modeling, we found that the interaction data (described in Subsection \ref{EachInteraction} and in the bottom part of Table \ref{tab:MeasuresQuaire}) alone were not yielding effective models. So to explore whether user attributes might account for some of the variability in \textbf{IntUsat}, we added demographic attributes gathered by the Recruitment questionnaire (see Subsection \ref{RecruitmentQuestionnaire} and top part of Table \ref{tab:MeasuresQuaire}), with some recoding of the categorical data into two categories. 

The MLR  was run excluding missing values listwise. This meant that starting with 136 full interactions, dropping the 8 without recruitment consent, those not identifying as Male/Female re-coded as missing \textbf{RecruitGender} values, and \textbf{IntCoffee-Correct} (Walked-Away, Ask-Repeat) values being re-coded as missing, 112 interactions remained and  were included in the MLR, the results of which are reported in Subsection \ref{SubSec: Quantitative Results}.

\subsection{Qualitative Analysis}
\label{subsec:QualitativeAnalysisMethod}
The qualitative questionnaire items were collected. A codebook was created 
using an inductive approach similar to grounded theory and open coding  \cite{McDonald19}. 
Two coders met to discuss the codebook, and unitisation policy \cite{campbell2013coding}. They then independently coded all the data. 
The average agreement across all codes was 95.8\% with a Cohen’s Kappa of 0.77. Themes were developed based on the coding \cite{Corbin08}. The themes and contributing codes are described in the next section along with the qualitative results. 

\section{Results and Discussion}
\label{Sec: Result and Discussion}
Firstly,  we examine quantitative results, addressing their bearing on \textbf{RQ1} and \textbf{RQ2}, then the qualitative results, as they related to \textbf{RQ1} alone, and finally we focus on \textbf{RQ3} and lessons learned.
\begin{table}[th]
\centering
  \caption{Significant Correlations: Showing each bi-variate correlation, with Pearson's \textit{r} and two-tailed \textit{p}-values by RQ.}
  \label{tab:CorrelationResults}
        \begin{tabular}{{c l l c c}}\toprule
    \textbf{Row No} &\textbf{RQ} & \textbf{Variable pair }&  \textbf{\textit{r}} & \textbf{\textit{p}}\\ 
    \midrule
    \small 1 & 1 & ExitTrust v. ExitUsefulness & .765 & $<$.001\\ 
    \small 2 & 1 & ExitTrust v. ExitAdaptability & .616 & .003\\ 
    \small 3 & 1 & ExitTrust v. ExitLikeabilty  & .535 & .012\\
    \midrule
    \small 4 & 1 & RecruitPTT v. ExitUsefulness & .600 & .004\\
    \small 5 & 1 & RecruitPTT v. ExitAdaptability & .569 & .007\\
    \midrule
    \small 6 & 1 & IntMeanUsat v. ExitUsefulness  & .567 & .009\\
    \small 7 & 1 & IntMeanUsat v. ExitAdaptability  & .547 & .013\\
    \midrule
    \small 8 & 2 & RecruitNARS v. ExitNARS  & .699 & $<$.001\\
    \small 9 & 2 & RecruitPTT v. ExitNARS  &  \textbf{-}.631 & .002\\
   \bottomrule
  \end{tabular}
\end{table} 
\subsection{Quantitative Results}
\label{SubSec: Quantitative Results}
\textbf{RQ1:} What factors play a role in the long-term user perceptions of the Robo-Barista?

As illustrated in Table \ref{tab:CorrelationResults}, trust was a clear factor. Specifically, there were strong\footnote{Pearson's \textit{r} of 0.5 and over equates to ``strong", while 0.3 to 0.5 equates to ``medium" 
\cite{Field09}} positive correlations between the trust in the robot after the final interaction (\textbf{ExitTrust}) and perceived usefulness (\textbf{ExitUsefulness}), adaptability (\textbf{ExitAdaptability}) and general likeability (\textbf{ExitLikeabilty}). (Table \ref{tab:CorrelationResults} rows 1 to 3). This is evidence that trust is either influenced by or is influencing views on those three acceptance measures. There were also strong positive correlations between users' propensity to trust (\textbf{RecruitPTT}) and both usefulness and adaptability (Table \ref{tab:CorrelationResults} rows 4 and 5).  This is another indication that trust is an active factor here. 


Additionally, there was a strong positive correlation between user satisfaction (\textbf{IntMeanUsat}) and both \textbf{ExitUsefulness} and \textbf{ExitAdaptability} (Table \ref{tab:CorrelationResults} rows 6 and 7), which was not surprising as prior work found that users' perception of the robot contributes to their interaction satisfaction \cite{Rossi21}.

\begin{mdframed}
\textbf{Finding 1: Trust is important in perceptions of the Robo-Barista}
Perceived trust of the robot and an individual's propensity to trust affect the perceptions of usefulness, adaptability and likeability of the robot.

\end{mdframed}

MLR analysis allowed a deep dive into factors that affected user satisfaction. Table \ref{tab:RegressionResults} gives the variables used in this model
 in order of influence, as given by the Standardized Beta coefficient. The model has $R^2 = .339$, which indicates that  33.9\% of the variability in \textbf{IntUsat} is explained by these seven variables. 

\begin{table}[ht]
\centering
  \caption{The Multiple Linear Regression results. The influence on IntUsat of each variable in the Model as revealed by their Standardized Beta coefficient.}
  \label{tab:RegressionResults}
        \begin{tabular}{{l c}}\toprule
    \textbf{Variable} & \textbf{Standardized Beta coefficient} \\ 
    \midrule
    \small IntCoffee-Correct & .270\\
    \small RecruitGender & .251 \\
    \small RecruitAge & -.232 \\
    \small RecruitNARS & -.210\\
    \small RecruitCoffeeHabit &  -.210\\
    \small IntTurns-user & -.166\\
    \small IntTime-taken & -.116 \\

   \bottomrule
  \end{tabular}
\end{table}

 Table \ref{tab:RegressionResults} can be interpreted as follows: \textbf{IntUsat} was higher: when a user identified as female (\textbf{RecruitGender}), when  users' were younger (\textbf{RecruitAge}), when their negative attitude to robots was lower (\textbf{RecruitNARS}), when the user's coffee habit was low (\textbf{RecruitCoffeeHabit}), when the robot dispensed the correct coffee (\textbf{IntCoffee-Correct}), when they took fewer turns in an interaction (\textbf{IntTurns-user}), and when their interaction was shorter (\textbf{IntTime-taken}).

\begin{mdframed}
\textbf{Finding 2: Interaction features and user attributes predict user satisfaction }
Gender, age, attitude to robots, and their coffee habits all contribute to user satisfaction, as well as, whether they got the right coffee and how long it took.

\end{mdframed}

\textbf{RQ2:} Did user attitude towards robots change over the study period?

Given that we had measured \textbf{NARS} at recruitment and exit, we were able to compare users' negative attitudes to robots at the the start and end of the study. A paired sample t-test, (\textbf{RecruitNARS} Mean, 2.39, \textbf{ExitNARS} Mean, 2.40, N=21, t(20)=$-$.122, sig two-tailed p=0.904) revealed no statistically significant difference between the two measures. This shows that users' negative attitudes to robots did not change through the study, confirming \textbf{NARS} as a relatively stable characteristic measure. By way of confirming the t-test result, there is also a strong positive correlation of \textbf{RecruitNARS} with \textbf{ExitNARS} (Table \ref{tab:CorrelationResults} row 8).

We also report here the strong negative correlation between Propensity to Trust (\textbf{RecruitPTT}) and 
users' negative attitude to robots after completing the experiment (\textbf{ExitNARS}) (Table \ref{tab:CorrelationResults} row 9). This confirms users with low Propensity to Trust are also likely to have a high negative attitude to robots, as seen in prior work \cite{lim2022we}.

\begin{mdframed}
\textbf{Finding 3: Negative attitudes to robots did not change during the study}
\textbf{NARS} did not change from 
start to 
end of the study, reflecting that \textbf{NARS} is a reliable measure of a stable trait, confirming prior work. 
\end{mdframed}

\subsection{Qualitative Results}

\textbf{Returning to RQ1:} What factors play a role in the long-term user perceptions of the Robo-Barista?

The views expressed in the qualitative responses of the 21 \textbf{Thru-Study} users (mostly yielded from the Exit questionnaire) give some interesting insights into how users viewed the Robo-Barista and contribute to our understanding of \textbf{RQ1}. The one closed question in the `Exit questionnaire' - ``What is the Robo-Barista to you?" - was overwhelmingly answered by choosing ``A tool to perform tasks" (N=18) with 2 users answering ``A social agent" and one, ``A social mediator among common room visitors". Although this mostly utilitarian view is reflected in the analysis of the open responses below with the majority focusing on practical aspects of the Robo-Barista service, a proportion of the views did focus on aspects such as the Robot's appearance, the quality of the Robot's conversation, and their enjoyment of the experience.

The analysis (described in Subsection \ref{subsec:QualitativeAnalysisMethod}) led to the development of themes expressed by the users. In our description of those views, we also indicate the percentage of the total 21 \textbf{Thru-Study} users who express them. The most popular themes are summarised in Table \ref{tab:themes}.

\begin{table}[htbp]
 \caption{Themes expressed and their percentage popularity among the 21 thru-study users (N).}
 \label{tab:themes}
    \begin{tabularx}{1.\linewidth}{p{.25\linewidth} p{.45\linewidth} c}
    \hline
    \textbf{Theme} & \textbf{Example codes }& \textbf{Expressed}\\
    & & \textbf{by \% of N}\\
    \hline
    Dialogue & Repetition, Conversation flow, Pitch & 86\%\\
    Coffee & Free, Selection, Quality & 67\%\\
    Feelings & Interest, Enjoyment, Boredom, Comfort &  62\%\\
    Service & Reliability, Benefits &  57\%\\
    Physical Interface & Cup placement, ID card reading& 52\%\\
    Pace & Duration, Speed & 38\%\\
    Being Understood & Accents, Background noise & 29\%\\
    Robot & Appearance, Demeanor & 19\%  \\
    \hline
    \end{tabularx}
\end{table} 

Users were asked whether or not they would continue using the Robo-Barista. The majority, 71\%, said they would, 24\% said they wouldn't and one (5\%) gave a conditional view ``\textit{Probably not, unless the coffee remains free}" [P9]. Indeed 24\% mentioned  free coffee being a motivating factor. Other reasons expressed for wishing to continue using the service included coffee quality, and the enjoyment and interest of the experience. 

Beyond the majority stating they viewed the robot as a tool, other indications of this utilitarian view are that, although the robot introduced itself as ``Alex", only two (10\%) referred to it by name in their feedback responses and only two made any comparison with human baristas.

The ``\textit{Dialogue}" theme, perhaps unsurprisingly with a conversational robot, was the most popular. Users wished for more flexible and free flowing chat (14\%) from the Robo-Barista and less repetition from one interaction to the next  (29\%). Some found the interaction initially challenging but it seems they may have adapted, something expressed by P11: ``\textit{I have adapted to the robot. I know where to stand ... how to speak to it}". This finding is inline with prior work \cite{Huttenrauch02} stating that users will over time adapt to and come up with workarounds for possible shortcomings to fulfill their needs. 

This leads on to another aspect of the Robo-Barista service in that for the duration of the study it was a feature of the common room, and it seems that others perhaps found it an entertaining feature beyond its coffee function. Here one user's feedback is perhaps alluding to the idiosyncrasies of the Robot's conversations when overheard from outside the booth: ``\textit{I enjoyed the interaction, the coffee, and had a good laugh hearing Alex talk to its customers"} [P19]. Some feedback mentioned the tiredness detector feature 
which was greeted both positively and negatively: ``\textit{I loved the suggestion to add caffeine too, it was quite funny}" [P17] and "\textit{Always gives the same suggestions for extra dose, do i constantly look tired ?"} [P4].  

Users could have a nuanced view of the robot e.g. P17 (quoted above)
began their answer stating, ``\textit{I was quite impressed by what it could do and understand.}", but when describing their desired feature wrote ``\textit{Add some functionality to make it shut up :D as harsh as that sounds, sometimes you just want to have a coffee}". Another new functionality suggestion was for visual display to show what the robot was seeing, which when taken in conjunction with one user expressing doubt as to whether the tiredness detector was simply random [P15], we can see that more visual feedback could perhaps help with transparency.
\subsection{Lessons Learned}
\label{Sec: Lessons Learned}
\textbf{RQ3:} What lessons are learned for a long-term study in the wild?

Throughout this paper, we have presented a number of lessons learned. 
Other challenges were mostly around logistics, some of which were expected (e.g. booting up at the start of day
), but there were also 
unforeseen technical challenges (e.g. the coffee machine doing unexpected cleans). These had to be met by the four-person on-site support team, 
which included two departmental support staff and two researchers, one taking the role of ``robot wrangler" \cite{TakayamaWrangler}, while the other fielded user emails and coordinated support. 

The Robo-Barista needed routine check-ups at the start and the end of each day, to clean, provision the coffee machine, to boot up and shut down the system. This duty was shared among the team on a rota. Aside from the challenge of coping with the sometimes noisy environment around the booth the most intractable technical challenge was in the unpredictable nature of the coffee machine's cleaning activity, a 20 second automated action. If the coffee machine initiated this rinsing procedure during a user interaction, it would cause the machine to report a ``service unavailable" error to the system. This would a) prevent the system properly responding to user requests and b) cause the system to generate an email alert to the support team. Unfortunately, there was no way to predict or control this behaviour of the coffee machine.

\begin{mdframed}
\textbf{Lesson 4: Expect technical and support challenges: }
 Supporting a robot installation ``In the Wild" may require more time and resources than you originally think might be needed.
\end{mdframed}

The ``Feelings and Dialogue-Interface" themes revealed that users often wish for a service that would adapt more to their individual use and offer more variety in interaction. e.g. ``\textit{I do not need the robot to tell me, every day, where to put my loyalty card.}"[P11], and ``\textit{Chat should perhaps be a bit more flexible. Having to go through the same questions again with it feels redundant}" [P6]. Indeed, the MLR results indicate that user attributes, including age and gender, influence \textbf{IntUsat} and thus show that this would be a good direction to explore further.
\begin{mdframed}
\textbf{Lesson 5: Flexibility, adaptability and variety in long-term study}
Aim to offer more flexibility, adaptability and variety in the robot interactions, as previous research suggested \cite{Rossi21, Langer19, Leite13}, with the possibility of developing interaction features to appeal to different user groups.

\end{mdframed}

\section{Conclusion and Future Work}

This paper presents an exploratory longitudinal ``In the Wild" study of social robot acceptance to better understand users’ perceptions of such robots over time, which can help to better shape future design and implementation. It has uncovered several important lessons and findings. The unexpected challenges occurring at various stages of the deployment perhaps is not surprising due to the uncertainties in the environment and the heterogeneous group of users \cite{Leite13}. Additional effort in terms of usability, flexibility and adaptation will be necessary in future studies. 
Ethics approval for a less formal engagement of users will be useful in such installation. 
In terms of findings, trust has been shown to play a key role in influencing acceptance of the robot. It is also interesting to see that users' satisfaction with the robot depends on individual differences, confirming previous findings \cite{Lewis18,lim2022we}. 
Thus, personalisation should be an important part of interaction design for these types of long term studies.
\section{Acknowledgements}
We would like to thank Jose David Aguas Lopes for his work on the original Robo-Barista. Special thanks go to Claire Porter and Jennifer Hurley for their indispensable support with the maintenance and upkeep of the Robo-Barista
. This work was funded and supported by the UKRI Node on Trust (EP/V026682/1) \url{https://trust.tas.ac.uk} and 
EPSRC CDT on Robotics and Autonomous Systems (EP/S023208/1).

\bibliographystyle{IEEEtran} 
\bibliography{IEEEabrv, main}
\end{document}